\documentclass[sn-mathphys]{sn-jnl}

\jyear{2022}%
\usepackage{subfigure}
\theoremstyle{thmstyleone}%
\theoremstyle{thmstyletwo}%

\theoremstyle{thmstylethree}%

\begin{document}

\title[Article Title]{Prompt-tuning for Clickbait Detection via Text Summarization}


\author[1]{\fnm{Haoxiang} \sur{Deng}}\email{mlxxwlj@126.com}
\author[1]{\fnm{Yi} \sur{Zhu}}\email{zhuyi@yzu.edu.cn}
\author[1]{\fnm{Ye } \sur{Wang}}\email{wangye\_lj@163.com}
\author[1]{\fnm{Jipeng} \sur{Qiang}}\email{jpqiang@yzu.edu.cn}
\author[1]{\fnm{Yunhao} \sur{Yuan}}\email{yhyuan@yzu.edu.cn}
\author[1]{\fnm{Yun} \sur{Li}}\email{liyun@yzu.edu.cn}
\author*[2]{\fnm{Runmei} \sur{Zhang}}\email{zrm55326@163.com}

\affil[1]{\orgdiv{School of Information Engineering}, \orgname{Yangzhou University}, \orgaddress{\street{88 South Daxue Road}, \city{Jiangsu}, \postcode{225127}, \country{China}}}
\affil[2]{\orgdiv{School of Electronics and Information Engineering}, \orgname{Anhui Jianzhu University}, \orgaddress{\street{292 Ziyun Road}, \city{Anhui}, \postcode{230022}, \country{China}}}


\abstract{Clickbaits are surprising social posts or deceptive news headlines that attempt to lure users for more clicks, which have posted at unprecedented rates for more profit or commercial revenue. The spread of clickbait has significant negative impacts on the users, which brings users misleading or even click-jacking attacks. Different from fake news, the crucial problem in clickbait detection is determining whether the headline matches the corresponding content. Most existing methods compute the semantic similarity between the headlines and contents for detecting clickbait. However, due to significant differences in length and semantic features between headlines and contents, directly calculating semantic similarity is often difficult to summarize the relationship between them. To address this problem, we propose a prompt-tuning method for clickbait detection via text summarization in this paper, text summarization is introduced to summarize the contents, and clickbait detection is performed based on the similarity between the generated summary and the contents. Specifically, we first introduce a two-stage text summarization model to produce high-quality news summaries based on pre-trained language models, and then both the headlines and new generated summaries are incorporated as the inputs for prompt-tuning. Additionally, a variety of strategies are conducted to incorporate external knowledge for improving the performance of clickbait detection. The extensive experiments on well-known clickbait detection datasets demonstrate that our method achieved state-of-the-art performance.}




\maketitle

\section{Introduction}\label{sec1}

Online services have provoked a vast amount of attention and research in recent decades. Some popular online applications, such as social media and news portal, have emerged as pervasive tools in daily life. With the rapid development of these online contents, more clicks and web traffic refer to more profit and commercial revenue \cite{qian2021hierarchical,R2naeem2020deep}. To this end, many publishers deliberately produce gimmicky and sensational headlines that do not match the actual content to lure users' attention or curiosity for increasing clicks, which is known as clickbait \cite{zheng2021deep}. The spread of clickbaits in online news, posts, and updates inevitably leads to dissatisfaction and even nasty of users. The clickbaits not only contain inaccurate or even error information to mislead readers and spread fake news, but also may be used to perform click-jacking attacks by redirecting users to phishing websites for stealing personal information \cite{shu2018deep,zaharia2023opportunities}. With the ever-increasing size of clickbaits, there is an urgent demand for developing effective automatic clickbait detection methods.

The research paradigm of clickbait detection evolved from early feature engineering-based methods to neural networks and, more recently, into pre-trained language models like BERT, which also show superiority in this task. Early feature engineering methods extracted linguistic features such as word frequency and sentiment polarity for discriminative classification to detect clickbait \cite{a4biyani20168,a5wei2017learning}, which are restricted to learn abstract features. Recently, deep neural networks have been widely applied in clickbait detection for the advantages of powerful representations learning \cite{a6yoon2019detecting}, while these methods usually require large amounts of labeled data for model training. In the past few years, Pre-trained Language Models (PLMs) have shown to be extremely helpful in a wide range of NLP applications including clickbait detection \cite{a8indurthi2020predicting,a9yi2022clickbait}. However, these PLMs methods, such as BERT \cite{a21devlin2018bert} and RoBERTa \cite{a22liu2019roberta}, typically require auxiliary information like news content to fine-tune the model. The huge gap between pre-training and fine-tuning will prevent downstream tasks from fully utilizing pre-training knowledge.

More recently, inspired by GPT-3, prompt-tuning has been widely explored to enable better semantic modeling in many natural language processing tasks \cite{a10brown2020language}. In the prompt-tuning, clickbait detection is performed by reformulating tasks as natural language "prompts" and completing these prompts with pre-trained language models. For example, given the news with headline $x$ as: "If you have ever used Google Docs for anything important, you should know about this." and the corresponding content $c$ as "... A new Chrome extension lets people see every edit made on a Google Doc, potentially revealing CV lies as well as serial editers ...", the prompt-tuning method with manual template wraps it into "Article Title: $x$, Article Content: $c$, is this clickbait? [MASK]" and predicts the category word like "Clickbait" or "Not-Clickbait" to fill the "[MASK]" token. Compared with the previous fine-tuning PLMs methods, prompt-tuning requires no additional neural layer and shows excellent performance even in few-shot or zero-shot learning scenarios.

Different from fake news, the crucial problem in clickbait detection is determining whether the headline matches the corresponding content. Despite most existing prompt-tuning methods have achieved impressive performance on clickbait detection, they mainly compute the semantic similarity between the headlines and contents. However, the differences in length and semantic features between headlines and contents are always significant. For example, the headline is usually less than 20 words, and the length of news content is often dozens of times longer. To this end, directly calculating semantic similarity between headlines and content is often difficult to summarize the relationship between them.

In this paper, we propose an intuitive Prompt-tuning method for Clickbait detection via Text Summarization, short for PCTS. Text summarization is introduced to summarize the contents and clickbait detection is performed based on the similarity between the generated summary and the contents. For the problem of the huge gap between news headlines and contents, we exploit recent advances in the prompt-tuning model \cite{a11liu2021pretrain}, a two-stage text summarization model is introduced to produce high-quality news summaries, and both the headlines and new generated summary are incorporated as the inputs for prompt-tuning. Furthermore, five different strategies are conducted to incorporate external knowledge of clickbait detection, each of these strategies captures a different aspect of the characteristics of the expanded words, and all of them are integrated into the final verbalizer for prompt-tuning. The extensive experiments on well-known clickbait detection datasets demonstrate that our method achieved state-of-the-art performance. In summary, the contributions of this paper are as follows:

\begin{itemize}
	\item We propose an intuitive prompt-tuning method for clickbait detection via text summarization (PCTS), which can take full advantage of prompt-tuning to predict the masked label for clickbait detection.
	\item Different from the previous works, our method generates high-quality news summaries to avoid the huge gap between the news headlines and contents. Moreover, five different strategies are proposed to construct verbalizer for optimal prompts. All of these significantly improve detection performance.
	\item We experimentally show that the PCTS is more robust and effective than SOTA baselines on benchmark datasets for clickbait detection tasks.
\end{itemize}

\section{RELATED WORK}
\subsection{Clickbait Detection}

Clickbait detection has gained considerable attention due to the surge in clickbait prevalence across social media platforms. Accurate automatic detection methods are in high demand. Earlier approaches primarily relied on feature engineering, incorporating semantic \cite{a12}, linguistic \cite{a13blom2015click}, and multi-modal features \cite{a1chen2015misleading}. For instance, Prakhar et al. formalized the concept by defining eight types of clickbait and utilized features such as content, similarity, URL, informality, and forward reference for machine learning model training\cite{a4biyani20168}. Martin et al. proposed an automated clickbait detection approach utilizing 215 features, which were categorized into teaser messages, linked web pages, and meta information. Various classifiers and evaluation metrics were employed, with the random forest classifier exhibiting optimal performance in terms of ROCAUC, precision, and recall \cite{a14potthast2016clickbait}. In a different vein, Wei et al. categorized clickbait into ambiguous and misleading news. Ambiguous news detection involved mining class sequential rules as features, combined with basic features for classifier training. For misleading news detection, consistency features between headlines and news bodies were derived, employing a semi-supervised co-training approach for classification \cite{a5wei2017learning}. However, these methods necessitated specialized expertise for feature identification and manually crafted features encountered limitations in effectively capturing semantic information.

With the advent of deep neural networks, various deep learning approaches have been employed for clickbait detection, showcasing promising outcomes. These methods encompass Recurrent Neural Networks (RNN) \cite{a15anand2017we}, Convolutional Neural Networks (CNN) \cite{a16agrawal2016clickbait}, Attention Mechanism \cite{a17mishra2020musem}, and Graph Attention Networks \cite{a2liu2022clickbait}. For example, Agrawal et al. utilized word2vec to convert news headlines into word embeddings, proposing a CNN model for clickbait classification \cite{a16agrawal2016clickbait}. Zhou et al. introduced a novel self-attention mechanism at the token level of bidirectional Gated Recurrent Units (biGRU) and incorporated the Part of Speech Analysis Module (POSAM) employing Laplace Smoothing and NLTK POS Tagger. This not only reframed the task as a multi-classification problem but also enriched the model's linguistic insights \cite{R1zhou2017clickbait}. Rajapaksha et al. optimized the clickbait classification model by introducing new hierarchical structures and features, such as RNN and non-linear layers, along with a pruning strategy to reduce parameters and enhance operational efficiency \cite{R5rajapaksha2021bert}. Naeem et al. employed a deep learning framework, integrating the Part of Speech Analysis Module (POSAM) and the Long Short-Term Memory module (LSTM), utilizing a variant of the n-gram classifier with Laplace smoothing for part-of-speech analysis, and optimizing the LSTM model to achieve effective detection of clickbait in social media \cite{R2naeem2020deep}. Razaque et al. proposed the Legitimate and Illegitimate List Search (LILS) algorithm and the Domain Rating Check (DRC) algorithm, clickbait security integrates a deep recurrent neural network (RNN) featuring a gated recurrent unit (GRU) as an autoencoder. This augmentation significantly enhances the precision and security evaluation of clickbait, proficiently identifying covert malicious content within hyperlinks\cite{R23razaque2022clickbait}. Keller et al. introduced a three-branch system employing both shallow learned models and transformer-based approaches for spoiler type classification, with transformer models exhibiting superior performance. They also utilized a transformer-based question-answering model for spoiler creation \cite{R6keller2023nancy}. Despite these advances, the application of deep learning methods is hampered by the cost of acquiring labeled data. To address these challenges, methodologies like data augmentation \cite{a6yoon2019detecting} and domain adaptation \cite{a20lopez2018hybridizing} have been proposed, though the introduction of augmented and transferred features may introduce additional noise.

Recently, pre-trained language models (PLMs) such as BERT \cite{a21devlin2018bert}, RoBERTa \cite{a22liu2019roberta}, and T5 \cite{a23raffel2020exploring} have emerged as powerful tools for language understanding and generation. These models capture syntactic \cite{a24goldberg2019assessing}, semantic \cite{a25ma2019universal}, and structural \cite{a26jawahar2019does} information in language. The comprehensive knowledge encoded in PLMs enhances downstream tasks, including clickbait detection, yielding superior results \cite{a8indurthi2020predicting}. For example, Marreddy et al. integrated traditional feature extraction methods, such as Bag of Words (BoW) and TF-IDF, manual features (structural and part-of-speech), and distributed word representations (Word2Vec, GloVe, FastText, Meta-Embeddings), alongside pre-trained models (ELMo, BERT, RoBERTa, ALBERT, and ELECTRA) for clickbait detection. This comprehensive integration of methodologies leverages the strengths of traditional feature extraction, manually crafted features, distributed word representations, and advanced pre-trained language models to enhance the performance of clickbait detection \cite{R4marreddy2021clickbait}. RoBERTa served as the shared encoder, and task-specific multi-layer perceptrons were added for classification. Liu et al. employed multiple features for clickbait detection on WeChat, using BERT, Bi-LSTM, and Graph Attention Network for semantic and syntactic information \cite{a2liu2022clickbait}. They treated clickbait detection as a ternary classification task, distinguishing between malicious-clickbait and general-clickbait. Hagen et al. proposed clickbait spoiling, utilizing RoBERTa for spoiler type classification and spoiling generation \cite{a28hagen2022clickbait}. Bilal et al. employed a methodology integrating Fine-tuned BERT with a neural network layer and conducted a comparative analysis of various text representation methods, including TF-IDF, ArabicBERT, RoBERTa, and TensorFlow's built-in embedding layer. In terms of model design, a comprehensive evaluation utilizing feedforward neural networks, LSTM, BiLSTM, and CNN among other deep learning models achieved optimal performance \cite{R22r2023transformer}. While fine-tuning PLMs has shown success, these methods often necessitate auxiliary information, such as news content, for effective fine-tuning. The substantial gap between pre-training and fine-tuning objectives poses challenges in leveraging the full knowledge potential within PLMs.

\subsection{Prompt-tuning}

The concept of prompt-tuning, as inspired by GPT-3 \cite{a10brown2020language}, has gained prominence as a method for transferring knowledge to downstream tasks using cloze-style objectives. This approach has demonstrated superior performance, particularly in few-shot learning scenarios \cite{a11liu2021pretrain}. The prompt-tuning involved the learning of soft prompts through back-propagation to condition a frozen language model, which allows only one trainable soft prompt to be appended to the input text.

The primary components include a template and a set of label words, where the template is a background description of the current task and the label words are the high-probability vocabulary predicted by PLMs in the current context. Various hand-crafted prompts, which are discretely specified prompts that remain constant during training, have been designed and applied across different tasks, including knowledge probing \cite{a29petroni2019language}, relation extraction \cite{a30han2022ptr}, and text classification \cite{a31hu2021knowledgeable}. For example, Gu et al. proposed the "Pre-trained Prompt Tuning" framework, addressing challenges in few-shot learning through the pre-training of soft prompts via self-supervised tasks on unlabeled data. Strategies such as verbalizer selection and real word initialization were introduced to overcome these challenges\cite{R8gu2021ppt}. Han et al. proposed the Prompt Tuning with Rules (PTR) method, which addresses the challenge of designing effective prompts for complex many-class classification tasks. PTR encodes prior knowledge into rules, designs sub-prompts based on these rules, and aggregates them to achieve a balanced tradeoff between efficiency and effectiveness. Experimental results on tasks such as relation classification, entity typing, and intent classification demonstrated that PTR outperforms traditional methods and baselines\cite{a30han2022ptr}.

To alleviate the time-consuming and labor-intensive task of prompt design, recent efforts have explored automatic prompt generation methods\cite{a32li2021prefix}. For example, Jiang et al. proposed a method combining mining-based and paraphrasing-based techniques for automatic prompt generation \cite{a33jiang2020can}. Shin et al. developed the AUTOPROMPT method, employing a set of trigger tokens determined through a gradient-based search \cite{a34shin2020autoprompt}. Zhou et al. introduced the Automatic Prompt Engineer (APE), a method designed to automatically steer large language models in generating precise and high-quality instructions. By leveraging iterative adaptive score estimation and employing two distinct generation strategies, APE effectively enhances the performance of natural language program synthesis tasks. It demonstrates substantial advantages, especially in areas such as zero-shot learning, few-shot learning, and chain-of-thought reasoning. Zhu et al. introduced Prompt-aligned Gradient (ProGrad) to address the issue of improper fine-tuning in large pre-trained vision-language models like CLIP. ProGrad aligns the gradient direction with pre-defined prompt predictions, preventing prompt tuning from forgetting general knowledge \cite{R11zhu2023automating}.

In the context of prompt-tuning, the verbalizer plays a crucial role as a projection from label words to categories, proving to be an important and effective strategy \cite{a36gao2020making}. While hand-crafted verbalizers have demonstrated sound performance in downstream NLP tasks, their reliance on prior knowledge may introduce omissions and biases in knowledge expansion. Consequently, automatic verbalizer construction methods in prompt-tuning have been proposed \cite{a37schick2020automatically}. For example, Hu et al. introduced a knowledgeable prompt-tuning method by integrating external knowledge into the verbalizer, selecting related words from the open Knowledge Graph, and refining them to filter out noise \cite{a31hu2021knowledgeable}. Wei et al. proposed a prototypical prompt verbalizer method, eliciting knowledge from pre-trained language models and learning prototypical embeddings for different labels in the latent feature space \cite{a38wei2022eliciting}. Jiang et al. introduced the MetaPrompter method which employs a meta-learning strategy by combining a prompt pool with an innovative soft Verbalizer (RepVerb) \cite{R13jiang2023effective}. RepVerb directly constructs label embeddings from feature embeddings, eliminating the need for learning additional parameters. Simultaneously, MetaPrompter utilizes a prompt pool with an attention mechanism to extract more task knowledge, generating instance-specific prompts. This approach enhances performance on complex tasks while mitigating computational and memory burdens. Chen et al. suggested adapting pre-trained language models to end tasks by considering both prompts and the verbalizer, utilizing PLMs to predict candidate words and a Natural Language Inference model for similarity measurement \cite{a39chen2022adaprompt}. Currently, methods that surpass the automatic generation of verbalizers have emerged. Cui et al. proposed ProtoVerb method learns prototype vectors directly from training data through contrastive learning, serving as Verbalizers for task prediction \cite{R12cui2022prototypical}. By optimizing the InfoNCE estimator objective, which involves maximizing intra-class instance similarity and minimizing inter-class instance similarity, the method proves significantly superior to automatic Verbalizers, especially demonstrating consistent improvements on untuned PLMs in limited data scenarios.

In practice, prompt-tuning methods have proven to be applicable to downstream tasks such as clickbait detection. For instance, Wu et al. introduced PEPL, a part-of-speech-enhanced prompt construction method, as an extension of prompt learning \cite{p1wu2023detecting}. This method utilizes part-of-speech tagging to identify syntactic feature words in titles. Subsequently, by constructing an augmented prompt, the model's attention is directed towards these syntactic feature words, thereby facilitating semantic comprehension guided by grammatical features. This innovative approach demonstrates its effectiveness in eliciting knowledge within pre-trained language models (PLMs). Thus, it is evident that through task reframing, manual template creation, and language expression, prompt-tuning indeed enhances the performance of clickbait detection in low-sample learning scenarios. However, current research applying prompt-based methods to clickbait detection remains relatively limited.

\subsection{Text Summarization}

Text summarization is a prominent and challenging area in Natural Language Processing (NLP) \cite{sum1widyassari2022review}. It involves the task of creating a concise and coherent summary of lengthy text documents, highlighting key points while maintaining the document's overall meaning. This process aims to produce summaries that efficiently capture the primary concepts, enabling a more efficient understanding of extensive textual information.

Two primary paradigms govern text summarization: extractive and abstractive. In the extractive approach, specific keywords from the input are chosen for output generation, aligning well with this model. Many extractive summarization methods treat the task as a sequence classification problem, utilizing diverse encoders such as recurrent neural networks \cite{sum12nallapati2016abstractive} and pre-trained language models \cite{sum6zhang2023diffusum}. Nallapati et al. introduced SummaRuNNer, utilizing a dual-layer bidirectional GRU-RNN for document extractive summarization \cite{t1nallapati2017summarunner}. They introduced an abstract training method that leverages Rouge score optimization to transform abstract summaries into extractive labels, enabling training exclusively based on human-generated summaries. Xie et al. introduce KeBioSum, utilizing SciBERT to identify PICO elements and incorporating domain knowledge into base PLMs through generative and discriminative adapters \cite{t2xie2022pre}. The method significantly improves the performance of extractive summarization in the biomedical domain. An alternative view considers extractive summarization as a node classification problem, utilizing graph neural networks to model inter-sentence dependencies \cite{sum10zhang2023contrastive}. Cui et al. present Topic-GraphSum, an extractive text summarization approach that integrates graph neural networks and neural topic models \cite{t3cui2020enhancing}. Document encoding utilizes BERT, and the neural topic model (NTM) learns document topics. The graph attention layer captures intersentence relationships, facilitating the efficient extraction of key information and addressing limitations in existing models regarding intersentence relationships and neglect of topic information.

In the realm of abstractive summarization, significant progress has been achieved through Seq2Seq models \cite{sum12nallapati2016abstractive}. These models build a neural network to establish a true relationship between input and output, refining the process with local attention mechanisms and advanced techniques like BERT. Architectures combining neural convolutional networks (CNN) with bidirectional LSTM showcase advancements in abstractive summarization, outperforming traditional methods. For example, Song et al. introduce the Abstractive Text Summarization (ATS) framework, which encompasses the extraction of semantic phrases from source text utilizing a phrase-based LSTM-CNN model \cite{t4song2019abstractive}. Subsequently, it generates summaries through a sequence-to-sequence model. Featuring a dual-decoder mechanism involving generate and copy modes, the model predicts the next phrase by encoding annotations in the source text. This approach notably improves the model's capability to construct new sentences, addressing both semantic and syntactic aspects. Recent advancements in abstractive summarization highlight the efficacy of diverse abstract generation, re-ranking strategies, and innovative architectures for second-stage summarization in enhancing overall summary quality and relevance \cite{tishenravaut2022summareranker}. Addressing the "exposure bias" associated with standard maximum likelihood teacher forcing is crucial for improving summary quality. Liu et al. introduce a second-stage summarization framework to tackle train-test distribution mismatch \cite{sum3liu2021refsum}. Transformer-based models like GSum and BERTSum, independently trained on varied data percentages, generate multiple candidates through diverse beam search during inference. A binary classifier, trained on GSum's outputs, selects the optimal summary, achieving state-of-the-art ROUGE scores on CNN/DailyMail. Furthermore, Liu et al. adopt a contrastive learning approach for second-stage summarization, demonstrating state-of-the-art performance on CNN/DailyMail through re-ranking BART-generated summaries \cite{sum4liu2021simcls}. Ravaut et al. present SummaReranker, an innovative multi-task mixture-of-experts architecture for re-ranking \cite{tishenravaut2022summareranker}. Surpassing strong baseline models across diverse datasets, including CNN/DailyMail, XSum, and Reddit TIFU, SummaReranker concatenates source documents and summary candidates. It utilizes a RoBERTa encoder and integrates a MoE layer for multiple prediction towers, optimizing various evaluation metrics. These advancements collectively contribute to the ongoing refinement of text summarization systems.

The increasing adoption of prompt-tuning and text summarization, as proposed in our study, represents a relatively unexplored domain within the clickbait detection literature. We introduce the SummaReranker model to generate high-quality news summaries, incorporating both the headlines and the summaries as input for the prompt-tuning model. This provides a distinctive perspective to enhance the performance of clickbait detection. Our approach significantly improves the ability to detect clickbait, particularly in scenarios involving small-sample learning.

\section{METHODOLOGY}

\subsection{Formalization and Overall Architecture}

Clickbait detection addresses a binary classification problem that involves determining whether news articles or social media posts contain clickbait content. Given the headline of news denoted as $x$, the corresponding content is denoted as $c$ with the binary label $y$, where $y = 1$ indicates the presence of clickbait, and $y = 0$ signifies otherwise. Since $x$ typically consists of a concise text, clickbait detection can be regarded as a short-text classification task, represented as $(y\mid x,c)$. The primary objective is to predict the label $y$ based on the textual information present in both the headline and content.

In contrast to existing methods, in this paper, we present an intuitive Prompt-tuning method for Clickbait detection via Text Summarization (PCTS). For the problem of the huge gap between news headlines and contents, we introduce the SummaReranker model to produce high-quality news summaries, incorporating both the headlines and the summaries as input for the prompt-tuning model. Moreover, we employed a variety of strategies to improve the performance of the model for clickbait detection. The overall architecture is illustrated in Figure 2. More specifically, our model consists of the following key components:

\begin{itemize}
	\item \textbf{Text Summarization:} The SummaReranker is a two-stage method that comprises summary candidate generation and re-ranking model based on evaluation metrics such as ROUGE-1, ROUGE-2, and ROUGE-L. For news summarization, we first utilize the PLMs to generate multiple summary candidates relevant to the contents of news. Subsequently, multiple evaluation metrics are employed to score the summary candidates for optimal summary selection.

	\item \textbf{Verbalizer Construction:} In the prompt-tuning model, verbalizer refers to the high-probability vocabulary predicted by PLMs in the current context, which has been proven to be effective in classification tasks \cite{a36gao2020making}. In our methods, five strategies, including Concepts Retrieval, BERT Prediction, FastText Similarity, Frequency-based Selection, and Contextual Information, are proposed to capture different characteristics of the expanded words for constructing verbalizer. These strategies can not only alleviate the negative effects of noise but also improve the performance of clickbait detection.

	\item \textbf{Clickbait Detection:} Following the construction of the verbalizer, the average predicted scores are utilized for clickbait detection. The method transforms the probability of clickbait into an objective function for the determination of the most suitable label. The parameters of the entire model are updated through the minimization of the cross-entropy loss function with L2 regularization to mitigate overfitting and enhance generalization.
\end{itemize}

\begin{figure}[!ht]
	\centering
	\includegraphics[width=\textwidth]{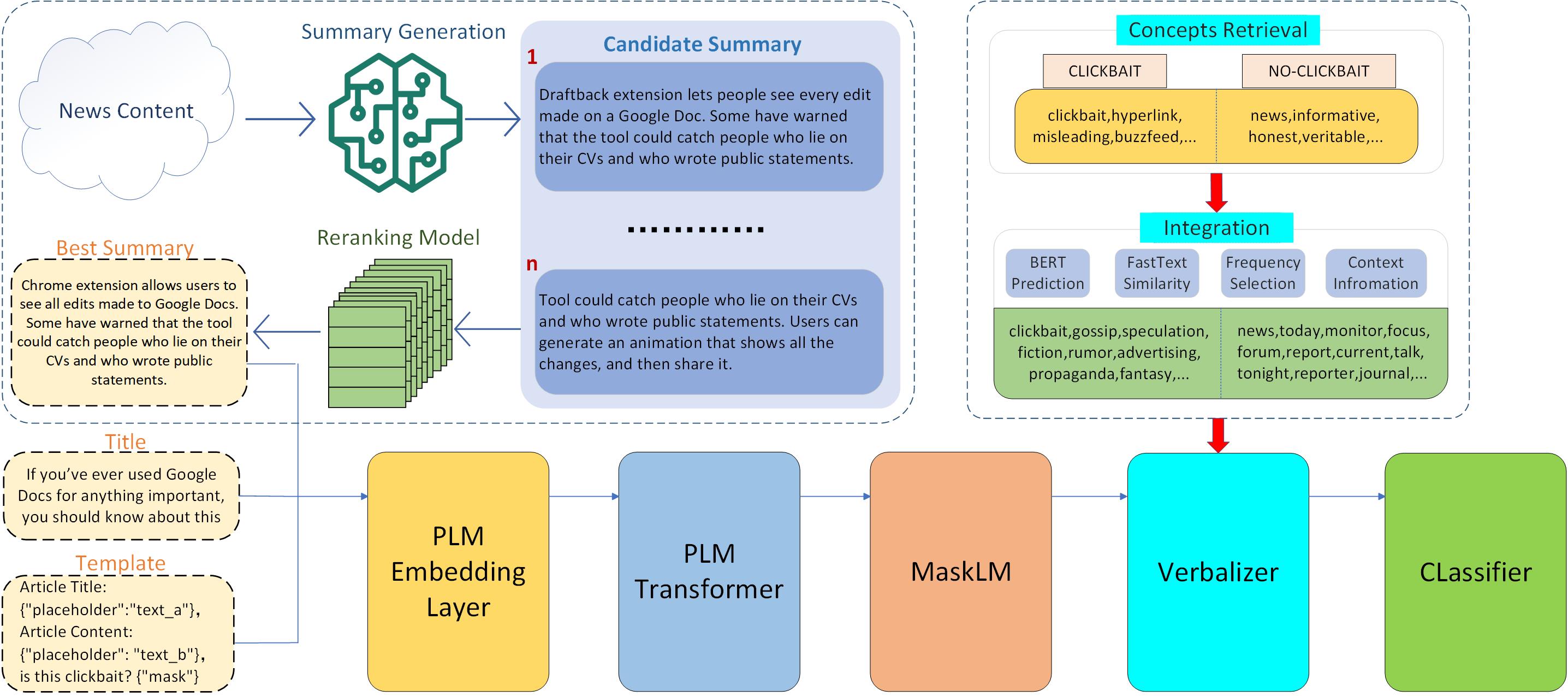}
	\caption{Illustration of our PCTS architecture. In the figure, the lower section is the core framework of our PCTS architecture, while the upper section provides a detailed view of the text summarization and verbalizer construction. The SummaReranker framework is employed to generate summaries, and through the implementation of five different strategies for integration, we facilitate the construction of the verbalizer. The final detection is conducted based on the headlines and new generated summaries by prompt-tuning.}
	\label{fig:your-figure-label}
\end{figure}

\subsection{Text Summarization}

\textbf{Summary Candidate Generation:} To generate high-quality summaries, we introduce \text{"pegasus-cnn\_dailymail"} as PLMs, and the PegasusTokenizer is utilized for text tokenization. For a given news content $c$, a series of candidate summaries are generated, which are represented as $PCS(c) = \{C_{1},..., C_{m}\}$, where $C_{m}$ represents the $m^{th}$ generated candidate summary.

\textbf{Re-ranking Framework:} After obtaining a set of candidate summaries, the task of re-ranking involves computing relevance scores for each candidate based on a set of evaluation metrics $M$, which includes three key evaluation metrics: ROUGE-1 (R-1), ROUGE-2 (R-2), and ROUGE-L (R-L).

\begin{itemize}
	\item ROUGE-1 (R-1): ROUGE-1 measures the unigram overlap between the generated text and reference text. It calculates the number of overlapping unigrams in the generated and reference text.
	\item ROUGE-2 (R-2): ROUGE-2 focuses on bigram overlap and evaluates the overlap of bigrams in the generated text and reference text.
	\item ROUGE-L (R-L): ROUGE-L measures overlap by finding the longest common subsequence between the generated text and reference text.
\end{itemize}

When considering an evaluation metric $\mu$ within a set of metrics $M$, we obtain associated scores for each candidate, represented as $S_{\mu} = \{\mu(C_{1}), \ldots, \mu(C_{m})\}$.Given $\mu(C_{m})$ as the evaluation score for the $m^{th}$  summary candidate, the optimal summary candidate $C_{u}^{*}$ can be identified as \eqref{Cm}:

\begin{equation}
	C_{u}^{*} = \arg \max_{C_{i} \in C}\{\mu(C_{1}),\ldots,\mu(C_{m})\}
	\label{Cm} \end{equation}

During the scoring of a candidate summary under various metrics $u$, it is possible for scores to be identical. Nevertheless, this does not impact the training process, where all candidate summaries are classified into two categories.
The problem is formulated as a binary classification task, where $C_{u}^{*}$ is considered the positive candidate, while all other candidates are treated as negative instances. The re-ranker is trained using binary cross-entropy loss as \eqref{Lmu}:

\begin{equation}
	L_{\mu} = -y_{i}\log(p_{\theta}^{u}(C_{i})) - (1-y_{i})\log(1-p_{\theta}^{u}(C_{i})) \hspace*{1em}
	\label{Lmu} \end{equation}
where $y_{i} = 1$ if $C_{i} = C_{u}^{*}$, otherwise $y_{i} = 0$.

To simultaneously optimize $N$ different metrics $M =\{\mu_{1}, . . . , \mu_{N} \}$, we employ a separate prediction head (tower) for each metric and minimize the average metric loss defined as \eqref{Lmini}:

\begin{equation}
	L = \frac{1}{N}\sum_{\mu \in M}L_{\mu} \hspace*{1em}
	\label{Lmini} \end{equation}
Each metric has its prediction head, and the model is trained to perform binary classification for each metric separately. The average loss $L$ measures how well the model performs across all the different metrics and is used to train the model to generate high-quality summaries considering multiple evaluation criteria simultaneously.

In our method, we use a pre-trained model checkpoint that is trained on candidate training with beam search and diverse beam search for the ROUGE-1, ROUGE-2, and ROUGE-L metrics to select the best summary for re-ranking.

\subsection{Verbalizer Construction}

In the prompt-tuning, the input statements are formalized as the natural language template, and the clickbait detection is regarded as the close-style task. For instance, in predicting the label $y$ of $x$: "If you have ever used Google Docs for anything important, you should know about this." and the summary $c$: "Chrome extension allows users to see all edits made to Google Docs. Some have warned that the tool could catch people who lie on their CVs and who wrote public statements." the template can be represented as:

\[
xp = [head] \; x, c:\text{In summary; it was} \; [MASK] \; : \; [tail]
\]

Given that the input $x = \{x_1, \ldots, x_n\}$ is classified into a category with label $y \in Y$, the label word set is denoted as $V_y = \{y_1, \ldots, y_n\}$, where $V_y$ is a subset of the whole vocabulary $V$, i.e., $V_y \subseteq Y$, and $V_y$ is mapped to a category with label $y$. In PLMs $M$, the probability that each word $v$ in $V_y$ is filled in the $[MASK]$ can be expressed as $p([MASK] = v \in V_y\mid x_{p})$. Thus, clickbait detection tasks can be framed as a probability calculation problem of label words, which can be computed as \eqref{ProCal}:

\begin{equation}
	p (y \in Y\mid x) = p ([MASK] = v \in V_{y}\mid x_{p}) \hspace*{1em}
\label{ProCal} \end{equation}

In the above example, if the calculated probability of $V_1 = \{clickbait\}$ for $y_1 = 1$ is higher than $V_2 = \{news\}$ for $y_2 = 0$, it indicates that the sentence $x$ is predicted as clickbait. In the context of automatic label word expansion or verbalizer construction, $V_y$ related to a specific category with label $y$ is expanded, such as $V_1 = \{clickbait\}$ can be extended as $V_1 = \{clickbait, hyperlink, misleading, \ldots\}$, leading to an evident improvement in clickbait detection performance, as validated in Section 4.6.

In the prompt-tuning model, verbalizer refers to the high-probability vocabulary in the current context. In our methods, five strategies are proposed to capture different characteristics of the expanded words for constructing verbalizer. The details are as follows:

\begin{itemize}
	\item Concepts Retrieval: To expand label word space for verbalizer in prompt-tuning, we first retrieve the related concepts from the open knowledge base. In the experiments, we introduce \textit{Probase} as an external knowledge base, which has specified the probability of each instance belonging to a concept and demonstrated robust performance in concept retrieval. Anchor words, such as the label names $y$ for clickbait and not-clickbait (e.g., news), are employed to retrieve $N$ concepts from \textit{Probase}, ranked by probabilities denoted as $N(v)$. As the number of retrieved concepts from \textit{Probase} is substantial, we compute the distance $dist(V_y, y)$ between each expanded label word set and each label name $y$ in the embedded space, initially refining the verbalizer. In experiments, for each input $x_i$, the top $N_i$ words are extracted whose vectors exhibit higher cosine similarity with the label name $y$. For all inputs, $\{N_1, \ldots, N_i, \ldots, N_n\}$ are ranked by similarity, and the top $N_a$ words are selected, excluding morphological derivations of $y$. Notably, in our experiments, we set $N_a$ to 15.
	\item BERT Prediction: Probability prediction is crucial for clickbait detection, providing context for the masked word in the template. PLMs are trained with two primary objectives: Masked Language Modeling (MLM) and Next Sentence Prediction (NSP). MLM focuses on predicting the "[MASK]" word in the template, chosen randomly from the input sentence. In our experiments, we used BERT to estimate the probability distribution of the vocabulary $p(\cdot\mid T  \backslash  [ MASK ])$ for the "[MASK]" word. Similar to concept retrieval, we selected the top $N_{a}$ words from this probability distribution. BERT prediction leverages contextual information in the input sentence to identify the most relevant words for categorization.
	\item FastText Similarity: Another widely used feature for constructing verbalizers is the similarity between concepts and the category name $y$. In addition to the semantic distance calculation used in concept retrieval, we utilize the fastText word embedding model \cite{joulin2016bag} to obtain vector representations of words. The similarity between the category name \( y \) and the expanded label words are measured by cosine similarity. Given the vector representations of the category name $y$ and expanded label word $s$ as $v_{y}$ and $v_{s}$, respectively, the cosine similarity is computed for similarity measurement. As with other strategies, words with lower similarity are discarded, and the top $N_{a}$ words are selected.
	\item Frequency-words selection: Frequency-words selection has been proven effective in creating verbalizers \cite{un}, is a common choice for improving classification performance in downstream tasks. Essentially, words with higher frequencies are more relevant in real-world contexts. Instead of using widely adopted corpora like Wikipedia, we employed Zipf frequencies from the SUBTLEX lists, which have demonstrated effectiveness in various NLP tasks. Furthermore, these word frequencies in the corpus have been found to correlate with human judgments of simplicity \cite{JipengQIANG,inproceedings}. $SUBTLEX^{3}$ comprises over six million sentences extracted from movie subtitles, and the Zipf frequency of a word is calculated as the base-10 logarithm of its occurrence per billion words. Similar to other strategies, we select the top $N_{a}$ words to match the desired output size.
	\item Context Information: In this strategy, we consider the contextual information surrounding the masked word. We utilize PLMs like BERT instead of traditional N-gram language modeling to capture contextual information. Given that BERT is a non-autoregressive language model, direct sentence likelihood computation is not feasible. To address this, we employ a symmetric window of size $c$ centered around the "[MASK]" word as context. Each word is masked from front to back and processed by the BERT model to compute the loss, and then we rank and select the top $N_{a}$ words based on their sequence loss. It's worth noting that in our experiments, we set $c$ to 5, and $N_{a}$ was fixed at 15 for all four strategies.
\end{itemize}

\subsection{Clickbait Detection}

After constructing the final verbalizer with five different strategies, we map the predicted probability of clickbait or not-clickbait to an objective function. In this process, each word in the final verbalizer is assumed to contribute equally to the prediction. Therefore, the average of the predicted scores is used for clickbait detection. This function can be calculated as \eqref{gCD}:

\begin{equation}
	g = \arg \max_{y \in Y} \frac{1}{\lvert V_{y}\rvert} \sum_{v \in V_{y}}p([MASK]=v  \mid  x_{p})\hspace*{1em}
\label{gCD} \end{equation}
where $\arg \max$ refers to the argument that maximizes the function, $Y$ represents the set of all possible labels, and $V_{y}$ represents the set of words in the final verbalizer for label $y$. The objective function $g$ can then be used to update the parameters of the entire model during the training process. Specifically, to update the parameters $\Theta$ of the entire model, we minimize the cross-entropy loss function \eqref{LossCD}:

\begin{equation}
	L =\frac{1}{N}\sum_{i=1}^{N}(y_{i}-p(y_{i}\mid x_{i}))^{2} +\frac{\lambda}{2k}||\Theta||^{2} \hspace*{1em}
\label{LossCD} \end{equation}
where the summation is taken over all training instances, $y_{i}$ represents the true label for the $i^{th}$ instance, $x_{i}$ is the input for the $i^{th}$ instance, and $\lambda$ is the coefficient of the L2 regularizer. The regularizer helps to prevent overfitting by penalizing large values of the model's coefficients, which can lead to better generalization performance.

\section{Experiments}\label{sec4}

In this section, we validated the effectiveness of our method. Firstly, we introduced the benchmark dataset and the implementation details of the experiments. Subsequently, we provided detailed experimental results and conducted an analysis. We then presented the influence of different templates. Finally, we conducted sensitivity and ablation experiments to assess the impact of various parameters.

\subsection{Datasets}

To evaluate the performance of our method, we conducted experiments on two well-known benchmarks: the News\_Clickbait dataset and the Webis Clickbait Corpus 2017 (Webis-Clickbait-17).
In addition to the posts, Webis-Clickbait-17 includes information about the articles linked in the posts. These posts were published from November 2016 to June 2017. To avoid publisher and topic bias, a maximum of 10 posts and publishers were sampled each day. All posts were annotated on a 4-point scale by five annotators from Amazon Mechanical Turk, and then the data text was preprocessed. For a news article to be classified as clickbait, it required an average score of more than 0.5 from the annotators. The statistical details of these datasets are presented in Table 1.

\begin{table}[!ht]
	\centering
	\setlength{\tabcolsep}{5pt}
	\caption{Statistics of the datasets. \# and avg.\# denote "the number of" and "the average number of".}
	\begin{tabular}{lll}
		\hline
		Statistics & News\_Clickbait & Webis-Clickbait-17 \\
		\hline
		\#total headlines & 28,423 & 38,517 \\
		\#train sets & 24,871 & 19,538 \\
		\#test sets & 3,552 & 18,979 \\
		avg.\#words per dataset & 499 & 634 \\
		avg.\#words per summary & 41 & 24 \\
		\hline
	\end{tabular}
\end{table}

\subsection{Baselines}

\begin{itemize}
	\item CNN \cite{a16agrawal2016clickbait}: Convolutional neural network is a typical deep neural network and extracted features from the text for clickbait detection.
	
	\item FastText \cite{joulin2016bag}: FastText is a text classification method that shares similarities with Word2Vec, which takes the sequence of words in the text into consideration. The learning algorithm employed by FastText bears a resemblance to the continuous bag-of-words (CBOW) model \cite{mikolov2013efficient}, which is a model used for learning distributed representations of words based on the order of words in the text.
	
	\item BERT \cite{a21devlin2018bert}: BERT is a bidirectional encoder-based representation method based on the Transformer architecture, which converts the clickbait detection task into a cloze-style question. By considering the contextual semantics of each word in the text, it effectively detects clickbait and serves as a baseline method.
	
	\item PL \cite{a10brown2020language}: The regular prompt-tuning involves filling the input sentence into manually crafted templates and forming a label word space using only category names.
	
	\item P-tuning \cite{a11liu2021pretrain}: This is an enhanced method based on prompt-tuning, which transforms template construction into a continuous parameter optimization problem.
	
	\item KPT \cite{a31hu2021knowledgeable}: This is an improved prompt-tuning method that expands the verbalizer with an external knowledge base. It is an optimized language prompt-tuning method that can more accurately detect clickbait.
	
	\item PEPL \cite{p1wu2023detecting}: PEPL integrated syntax-guided semantic comprehension into prompt-tuning. The method has introduced a critical component: the part-of-speech (PoS) feature, which can capture discrete semantic and grammatical attributes at the lexical level, especially when employed in Compositional Concept Detection (CCD) tasks.
\end{itemize}

\subsection{Implementation Details and Evaluation Metrics}

\subsubsection{Few-shot settings}
In the experiments, we randomly select $K$ (5, 10, 20) instances as the training set for PCTS and use the entire test set for validation. Recognizing that a very small number of training samples may significantly affect the effectiveness of the baselines, the training sample sizes are 100/200/400, corresponding to 5/10/20 shots, respectively. Considering that different choices of few-shot training samples and validation samples during the training process can impact the results, we replicate the same data with $K$ random seeds simultaneously. The number of training samples for the compared methods Bert, CNN, and Fast\_Text was set to 1000/2000/4000 for the News\_Clickbait dataset and 800/1600/3200 for the Webis-Clickbait-17 dataset, respectively.

\subsubsection{Parameter Settings}
Regarding CNN and Fast\_Text, we adopted the parameters of the original model, we encompassed a learning rate of 1e-5, a batch size of 16, and a maximum sequence length of 256.

For methods based on fine-tuning PLMs, such as BERT, we utilized the "bert-large-cased" version and conducted training with a batch size of 32 over 15 epochs.

For methods based on prompt-tuning, such as PL, P-tuning, KPT, and PEPL, we employed RoBERTabase \cite{a22liu2019roberta} as our PLMs. Our RoBERTa implementation is built upon the Hugging Face Transformer Library. During training, we set the hidden size to 768, the dropout rate to 0.5 to prevent overfitting, the learning rate to 4e-5, the batch size to 32, and the weight decay to 1e-5. Among all few-shot experiments, we established the number of epochs at 10 to ensure sufficient training and stable results. The Adam optimizer was used for model optimization.

All experimental results were conducted on a server with the following specifications: NVIDIA GeForce RTX 3090 Founders Edition, Intel(R) Core(TM) i9-10980XE CPU operating at a clock frequency of 3.00 GHz, and 125 GB of memory. For conducting the experiments, we used Python version 3.9.16 and pytorch-cuda version 11.7.

\subsubsection{Evaluation Metrics}
To measure the effectiveness of our detection method, we employ four evaluation metrics in our experiments:
\begin{enumerate}
	\item\textbf{Accuracy}: It represents the proportion of correctly predicted samples compared to the total number of samples.
	
	\item \textbf{Precision}: Precision is the ratio of the positive samples correctly predicted to the total positive samples predicted.
	
	\item \textbf{Recall}: Recall is the ratio of the positive samples correctly predicted to the total samples labeled as positive.
	
	\item \textbf{F1-score}: The F1-score is the harmonic average of precision and recall.
\end{enumerate}

\begin{table*}[!ht]
	\centering
	\renewcommand{\arraystretch}{1.1} 
	\caption{Results for accuracy (\%), precision (\%), recall (\%), and F1 score (\%) on \textbf{News\_Clickbait}. The values 100/5, 200/10, and 400/20 represent the training quantities for prompt adjustments (CNN, FastText, BERT and PL, P-tuning, KPT, PEPL, our PCTS). Bolder values indicate better performance.}
	
	\fontsize{10}{12}\selectfont 
	\begin{tabular}{lllllll}
		\hline
		\textbf{Shot} & \textbf{Method} & \textbf{Accuracy} & \textbf{Precision} & \textbf{Recall} & \textbf{F1-Score} \\
		\hline
		\multirow{6}{*}5
		& CNN & 0.5453 & 0.4597 & 0.6043 & 0.5222 \\
		& FastText & 0.5577 & 0.4352 & 0.2541 & 0.3209 \\
		& BERT & 0.5701 & 0.6270 & 0.5701 & 0.6101 \\
		& PL & 0.5393 & 0.7527 & 0.5393 & 0.5797 \\
		& P-tuning & 0.5089 & 0.7367 & 0.5089 & 0.5498 \\
		& KPT & 0.5663 & \textbf{0.7601} & 0.5663 & 0.6061 \\
		& PEPL & 0.5529 & 0.6718 & 0.5529 & 0.5940 \\
		& Ours & \textbf{0.5838} &0.7390 & \textbf{0.5838} & \textbf{0.6237} \\
		\hline
		\multirow{6}{*}{10}
		& CNN & 0.6153 & 0.5306 & 0.5590 & 0.5444 \\
		& FastText & 0.5936 & 0.5131 & 0.2280 & 0.3157 \\
		& BERT  & 0.6852 & 0.7434 & 0.6852 & 0.7069 \\
		& PL & 0.6994 & 0.7243 & 0.6994 & 0.7104 \\
		& P-tuning & 0.6078 & 0.7460 & 0.6079 & 0.6451 \\
		& KPT & 0.7054 & 0.7558 & 0.7054 & 0.7243 \\
		& PEPL & 0.7149 & \textbf{0.7622} & 0.7149 & 0.7326 \\
		& Ours & \textbf{0.7320} & 0.7420 & \textbf{0.7320} & \textbf{0.7367} \\
		\hline
		\multirow{6}{*}{20}
		& CNN & 0.6340 & 0.6307 & 0.2651 & 0.3733 \\
		& FastText & 0.6390 & 0.6578 & 0.2548 & 0.3673 \\
		& BERT  & 0.7930 & 0.7613 & 0.7930 & 0.7681 \\
		& PL & 0.7111 & 0.7223 & 0.7111 & 0.7164 \\
		& P-tuning & 0.7013 & 0.7526 & 0.7013 & 0.7205 \\
		& KPT & 0.7385 & 0.7477 & 0.7385 & 0.7428 \\
		& PEPL & 0.7561 & 0.7574 & 0.7561 & 0.7568 \\
		& Ours & \textbf{0.8003} & \textbf{0.7682} & \textbf{0.8003} & \textbf{0.7716} \\
		\hline
	\end{tabular}
\end{table*}

\begin{table*}[!ht]
	\centering
	\renewcommand{\arraystretch}{1.1} 
	\caption{Results for Accuracy (\%), Precision (\%), Recall (\%), and F1-scores (\%) on \textbf{Webis-Clickbait-17}. Others are the same as News\_Clickbait.}
	
	\fontsize{10}{12}\selectfont 
	\begin{tabular}{lllllll}
		\hline
		\textbf{Shot} & \textbf{Method} & \textbf{Accuracy} & \textbf{Precision} & \textbf{Recall} & \textbf{F1-Score} \\
		\hline
		\multirow{6}{*}5
		& CNN & 0.5488 & 0.2412 & 0.3981 & 0.3004 \\
		& FastText & 0.5839 & 0.2506 & 0.3565 & 0.2943 \\
		& BERT  & 0.5705 & 0.6290 & 0.5705 & 0.5937 \\
		& PL & 0.5766 & 0.6337 & 0.5766 & 0.5992 \\
		& P-tuning & 0.5884 & 0.6311 & 0.5884 & 0.6065 \\
		& KPT & 0.5904 & 0.6304 & 0.5904 & 0.6075 \\
		& PEPL & 0.5871 &\textbf{ 0.6338} & 0.5871 & 0.6065 \\
		& Ours & \textbf{0.6117} & 0.6328 & \textbf{0.6117} & \textbf{0.6215} \\
		\hline
		\multirow{6}{*}{10}
		& CNN & 0.6086 & 0.2485 & 0.3004 & 0.2720 \\
		& FastText & 0.6338 & 0.2420 & 0.2368 & 0.2394 \\
		& BERT  & 0.6690 & 0.6318 & 0.6690 & 0.6476 \\
		& PL & 0.6578 & 0.6301 & 0.6578 & 0.6424 \\
		& P-tuning & 0.6054 & 0.6307 & 0.6054 & 0.6168 \\
		& KPT & 0.6833 & 0.6345 & 0.6833 & 0.6538 \\
		& PEPL & 0.6140 & \textbf{0.6359} & 0.6140 & 0.6240 \\
		& Ours & \textbf{0.7047} &0.6323 & \textbf{0.7047} & \textbf{0.6568} \\
		\hline
		\multirow{6}{*}{20}
		& CNN & 0.6540 & 0.2541 & 0.2177 & 0.2345 \\
		& FastText & 0.6855 & 0.2482 & 0.1439 & 0.1822 \\
		& BERT  & 0.7103 & 0.6300 & 0.7103 & 0.6561 \\
		& PL & 0.7090 & 0.6333 & 0.7090 & 0.6580 \\
		& P-tuning & 0.7046 & 0.6308 & 0.7046 & 0.6559 \\
		& KPT & 0.7247 & \textbf{0.6349} & 0.7247 & \textbf{0.6595} \\
		& PEPL & 0.7024 & 0.6328 & 0.7024 & 0.6564 \\
		& Ours & \textbf{0.7566} & 0.5725 & \textbf{0.7566} & 0.6518 \\
		\hline
	\end{tabular}
\end{table*}

\subsection{Experimental Results}

The results on two different datasets are presented in Table 2 and Table 3. Despite the limited labeling of the headlines, our PCTS outperforms most of the baselines on the majority of the four metrics for both datasets. From the results, it can be observed that the methods based on prompt-tuning achieve better results compared to other baselines, which validated that the huge gap between rich knowledge distributed in the PLM and downstream tasks can be alleviated with prompt-tuning.

Furthermore, when the training data is limited, deep neural network methods without external prior knowledge perform worse than methods that incorporate external knowledge and pre-training information. We believe that in the absence of prior knowledge, the parameter updating process is susceptible to the limited distribution of training data, leading to poor generalization of the model.

In clickbait detection, news content contains a significant amount of redundant information. In our PCTS, we propose to generate summaries to assist detection, which can avoid the huge gap between the news headlines and contents. The results further confirm that training with news summaries and headlines as prompts is an effective and crucial strategy for mitigating text noise, extracting crucial features, and bridging the gap between label spaces. Due to the enhanced utilization of external knowledge and pre-training information in our method, it can achieve satisfactory detection results.

\subsection{Ablation Study}

In this section, we conducted an extensive series of ablation experiments across all datasets. These experiments were designed to discuss the distinct roles played by headlines and full news content for clickbait detection. In the experiments, we removed the text summarization (indicated as -summary) and the verbalizer construction (indicated as original news, which retained only the original category label vocabulary). The results are presented in Table 4, which demonstrates the model's superior efficiency in extracting critical information from articles when making predictions based on news summaries.

Training the model with news content that incorporates more crucial features aids in reducing the significant gap between news headlines and content. This improvement enables the model to better capture textual features, consequently enhancing its performance in the detection task.

\begin{table}[h]
	\centering
	\caption{The influence of different modules in the model on the results. The experiments are conducted with accuracy and F1 score over two datasets.}
	\begin{tabular}{llllllll}
		\toprule
		Datasets & Method & \multicolumn{2}{c}{shot:5} & \multicolumn{2}{c}{shot:10} & \multicolumn{2}{c}{shot:20} \\
		\cmidrule(lr){3-4} \cmidrule(lr){5-6} \cmidrule(lr){7-8}
		& & Acc & F1 & Acc & F1 & Acc & F1 \\
		\midrule
		\multirow{3}{*}{News\_Clickbait} & Ours & \textbf{0.5838} & \textbf{0.6237} & \textbf{0.7320} & \textbf{0.7367} & \textbf{0.8003} & \textbf{0.7716} \\
		& -summary & 0.5609 & 0.6013 & 0.6661 & 0.6900 & 0.7804 & 0.7501 \\
		& original news & 0.5637 & 0.6049 & 0.7191 & 0.7228 & 0.7575 & 0.7583 \\
		\midrule
		\multirow{3}{*}{Webis-Clickbait-17} & Ours & \textbf{0.6117} & \textbf{0.6215} & \textbf{0.7047} & \textbf{0.6568} & \textbf{0.7566} & 0.6518 \\
		& -summary & 0.5485 & 0.5775 & 0.6894 & 0.6560 & 0.7427 & 0.6573 \\
		& original news & 0.6087 & 0.6193 & 0.6826 & 0.6518 & 0.7299 & \textbf{0.6579} \\
		\bottomrule
	\end{tabular}
	
	\label{tab:mytable}
\end{table}

\subsection{Influence of Different Templates}

In prompt-tuning, although automatic template generation methods can alleviate the time-consuming and labor-intensive problems, most of these methods fail to achieve performance comparable to manual template methods in real-world scenarios \cite{a30han2022ptr}. Therefore, in our experiments, we conduct hand-crafted templates. We created four templates for clickbait detection tasks based on all four evaluation metrics on both two datasets. All the templates used in our experiments are listed in Table 5. For each template, we take the average of five experiments as the final result, which are all shown in Table 6.

\begin{table}[!ht]
	\centering
	\caption{Different Templates Used in Experiments}
	\setlength{\tabcolsep}{8pt} 
	\fontsize{10}{12}\selectfont 
	\begin{tabular}{|c|p{6cm}|}
		\hline
		\textbf{Id} & \textbf{Templates} \\
		\hline
		\multirow{5}{*}{1} & This title is: \{"placeholder": "text\_a"\} This article is \{"placeholder":"text\_b"\} and provides detailed information and analysis. \\
		& is\_clickbait: \{"mask"\} \\
		\hline
		\multirow{5}{*}{2} & This article title is: \{"placeholder": "text\_a"\}. The content is \{"placeholder": "text\_b"\} and provides detailed information and analysis. \\
		& Is it clickbait? \{'mask'\} \\
		\hline
		\multirow{3}{*}{3} & Article Title: \{"placeholder": "text\_a"\}, Article Content: \{"placeholder": "text\_b"\},
		is this clickbait? \{"mask"\} \\
		\hline
		\multirow{5}{*}{4} & This is an article about: \{"placeholder": "text\_a"\} This article discusses \{"placeholder": "text\_b"\} and provides detailed information and analysis. \\
		& is\_clickbait: \{"mask"\} \\
		\hline
	\end{tabular}
\end{table}

\begin{table}[!ht]
	\centering
	\renewcommand{\arraystretch}{1.2} 
	\caption{Comparison of Experimental Results on All Four Evaluation Metrics for Four Templates on the News\_Clickbait Dataset. The Best Results of Different Templates Are Highlighted in Bold.}
	\fontsize{10}{12}\selectfont 
	
	\begin{tabular}{lcccccc}
		\hline
		\textbf{Dataset} & \multicolumn{1}{c}{\textbf{T\_id}} & \multicolumn{4}{c}{\textbf{Metric}}
		\\\cmidrule{3-6} 
		
		& & \textbf{Acc} & \textbf{Pre} & \textbf{Rec} & \textbf{F1} \\
		\hline
		\multirow{4}{*}{News\_Clickbait} & 1 & 0.7336 & 0.7697 & 0.7336 & 0.7477 \\
		& 2 & 0.7938 & 0.7498 & 0.7938 & 0.7505 \\
		& 3 & \textbf{0.8003} & \textbf{0.7682} & \textbf{0.8003} & \textbf{0.7716} \\
		& 4 & 0.7260 & 0.7748 & 0.7260 & 0.7436 \\
		\hline
		\multirow{4}{*}{Webis-Clickbait-17} & 1 & 0.7137 & 0.6294 & 0.7137 & \textbf{0.6561} \\
		& 2 & 0.7386 & 0.6275 & 0.7386 & 0.6551 \\
		& 3 & \textbf{0.7566} & 0.5725 & \textbf{0.7566} & 0.6518 \\
		& 4 & 0.7488 & \textbf{0.6329} & 0.7488 & 0.6547 \\
		\hline
	\end{tabular}
\end{table}

\subsection{Parameter Sensitivity}

In this section, we will investigate the impact of parameters, including learning rate and batch size. When one parameter is changed, other parameters are fixed constant in the experiments. Based on the experimental results as shown in Figure 2, we observe that different datasets exhibit varying sensitivities to the learning rate. The learning rate determines the step size at which the model updates parameters during training, as well as whether the algorithm successfully converges and when it converges to a local minimum. Performance fluctuates significantly under different learning rates, which can be attributed to adjustments in the learning rate affecting the convergence speed of training.

Furthermore, the experimental results indicate that appropriately increasing the batch size can improve the detection performance across all datasets. This improvement can be attributed to larger batch sizes reducing gradient noise during training and effectively harnessing the parallel processing capabilities of GPUs.

\begin{figure}[!ht]
	\centering
	\includegraphics[width=0.45\textwidth]{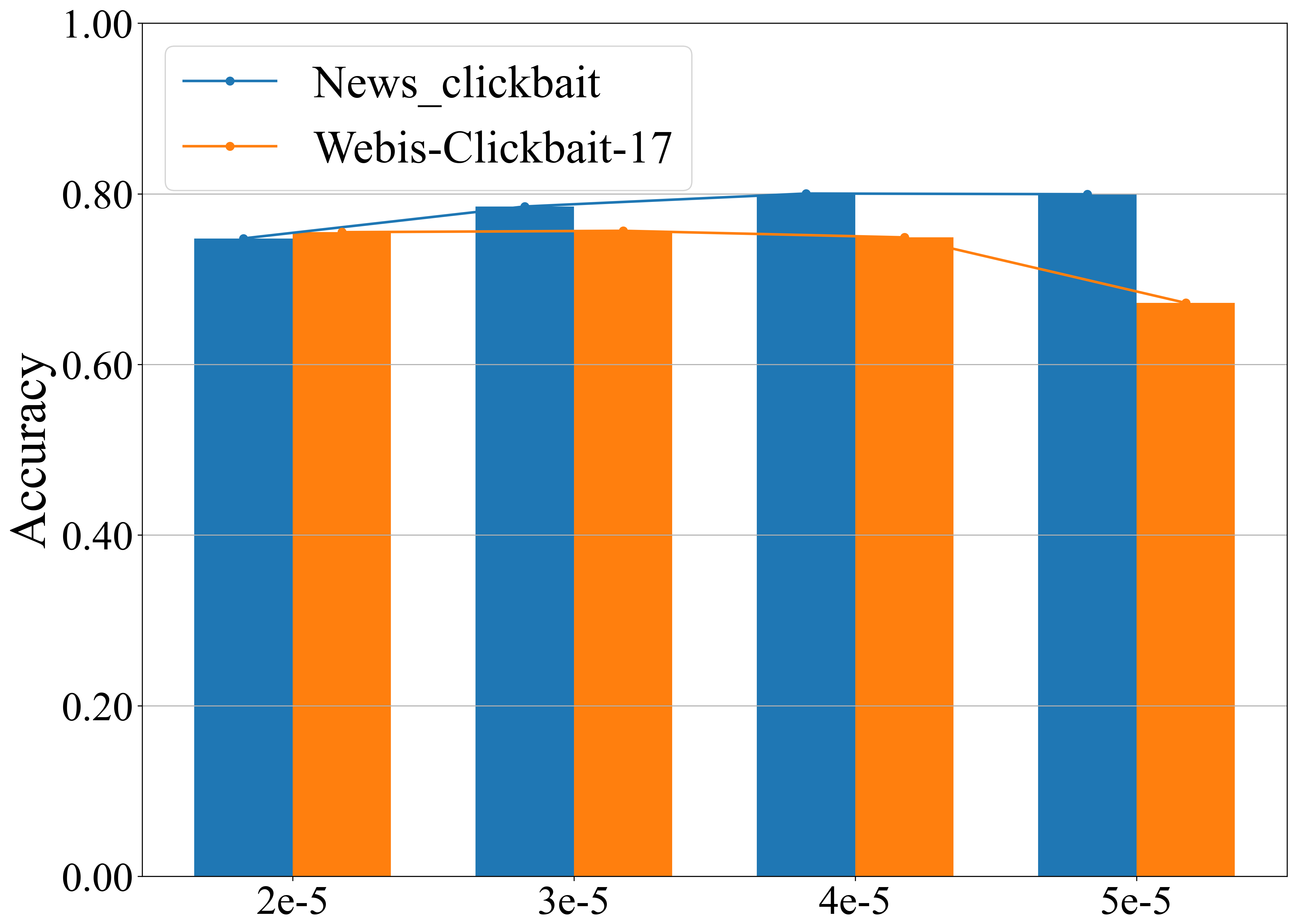}
	\caption{Learning Rate}
	\label{fig:learning-rate}
\end{figure}

\begin{figure}[!ht]
	\centering
	\includegraphics[width=0.45\textwidth]{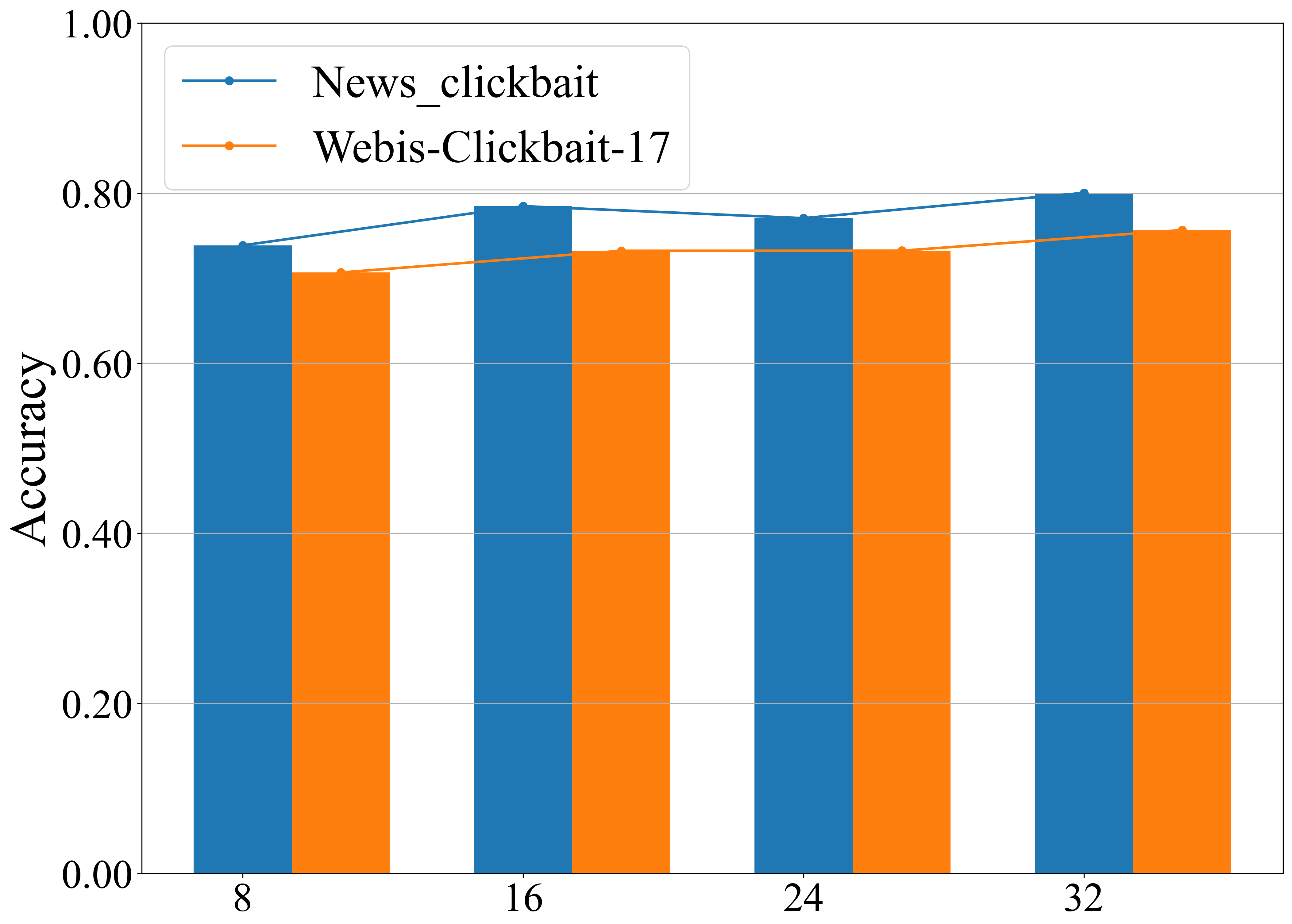}
	\caption{Batch Size}
	\label{fig:batch-size}
\end{figure}

\section{Conclusion}\label{sec5}

In this paper, we propose a prompt-tuning method for clickbait detection via Text Summarization. Text summarization is introduced to summarize the contents and clickbait detection is performed based on the similarity between the generated summary and the contents. Then five strategies are proposed to capture different characteristics of the expanded work for verbalizer construction. The experiments show the effectiveness of our method.

In the future, we will extend our research work in the following two directions. One is exploring better methods for automatic template construction on clickbait detection. The other is to incorporate more auxiliary information from external knowledge for some other tasks.

\section{Acknowledgements}\label{sec6}

This research is partially supported by the National Natural Science Foundation of China under grants (62076217), Yangzhou Science and Technology Plan Project City School Cooperation Special Project (YZ2023199), Open Project of Anhui Provincial Key Laboratory for Intelligent Manufacturing of Construction Machinery (IMCM-2023-01).

\section*{Declarations}

The authors declare that they have no known competing financial interests or personal relationships that could have appeared to influence the work reported in this paper.

\bibliography{sn-bibliography}

\end{document}